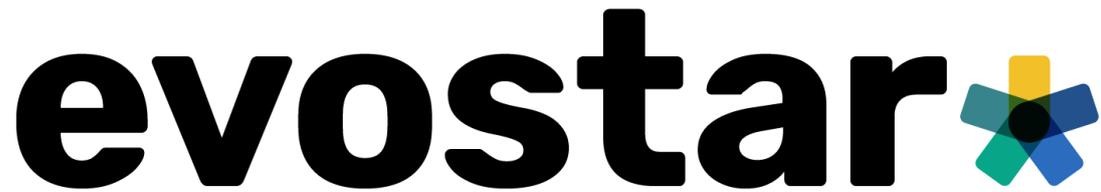

# Evo* 2022

## The Leading European Event on

## Bio-Inspired Computation

*Madrid (Spain). 20-22 April 2022*

# – LATE-BREAKING ABSTRACTS –

**Editors:**

**A.M. Mora**
**A.I. Esparcia-Alcázar**

# Preface

This volume contains the Late-Breaking Abstracts accepted at Evo* 2022 Conference, held in Madrid (Spain), from 20 to 22 of April.

They were also presented as short talks as well as at the conference's poster session.

The works present ongoing research and preliminary results investigating on the application of different approaches of Evolutionary Computation and other Nature-Inspired techniques to different problems, most of them real world ones.

These are very promising contributions, since they outline some of the incoming advances and applications in the area of nature-inspired methods, mainly Evolutionary Algorithms.

*Antonio M. Mora*
*Anna I. Esparcia-Alcázar*

# Table of contents

# Mapping the Field of Metaheuristic and Bioinspired Portfolio Optimization*

Feijoo Colomine Durán[1][0000−0002−2034−9205], Carlos
Cotta[2,3][0000−0001−8478−7549], and Antonio J.
Fernández-Leiva[2,3][0000−0002−5330−5217]

[1] Universidad Nacional Experimental del Táchira (UNET), Laboratorio de
Computación de Alto Rendimiento (LCAR), San Cristóbal, Venezuela
`fcolomin@unet.edu.ve`
[2] Dept. Lenguajes y Ciencias de la Computación, ETSI Informática, Universidad de
Málaga, Campus de Teatinos, 29071 - Málaga, Spain
`{ccottap,afdez}@lcc.uma.es`
[3] ITIS Software, Universidad de Málaga, Spain

**Abstract.** We analyze the bibliography related to portfolio optimization using metaheuristics and bioinspired algorithms. To this end, we perform data clustering based on lexical similarity between bibliographical descriptors and propose an internal arrangement of each cluster using evolutionary algorithms. We also conduct a network analysis in order to determine relevant keywords and their associations.

## 1   Introduction

Investment portfolios are standard tools based on the diversification of investments in several financial instruments or sectors, aimed to balance the expected return and the associated volatility of said investments, according to the degree of risk abhorrence of the investor. It is therefore an area which naturally lends itself towards multiobjective approaches. The richness of the problem (which can exhibit a number of additional constraints, e.g., cardinality restrictions, transaction lots, etc.) makes it generally untenable for resolution using simple approaches. Indeed, this problem has been extensively attacked with metaheuristic methods [2]. We hereby conduct a study of the field literature using data analysis techniques as well as evolutionary algorithms.

* This work is supported by Spanish Ministry of Economy under project DeepBio (TIN2017-85727-C4-1-P) and by Universidad de Málaga, Campus de Excelencia Internacional Andalucía Tech

## 2    Methodology

There are different approaches that can be used to analyze a research field, cf. [1,3]. Here, we have proceeded as follows: let $\mathcal{A} = \{A_1, \ldots, A_m\}$, be our raw input data, namely a collection of articles, where each $A_i$ is a bibliographical item described as a text string (title, abstract, keywords, etc.). In this case, this collection is obtained by querying the DBLP[4] in order to obtain articles featuring at least one term associated to the techniques of interest (namely, "evolutionary", "genetic", "swarm", "ACO", "metaheuristics", "tabu") and at least one term associated to the field under study (namely "portfolio", "investment", "markowitz", "sharpe"). Each $A_i$ is the title of a paper that fulfills these search criteria.

Each entry in the collection is subject to a stemming process so as to delete words that carry no semantic meaning, as well as to remove any syntactic inflections from the text. We thus obtain a collection $\mathcal{L} = \{L_1, \ldots, L_m\}$, where each $L_i$ is a set of meaningful stems associated to the $i$-th bibliographical item. Let $V = \cup_i L_i$ be the set of all stems used, and let $p = |V|$. $\mathcal{L}$ is therefore represented as a binary matrix $M_{m \times p}$, where the $i$-th row is the incidence vector of $L_i$ of $V$. This matrix will be used to cluster the data, using the cosine dissimilarity as distance measure [6], the k-means algorithm as clustering technique, and the silhouette method to determine the number of clusters measure [5]. To determine the significance of each stem within each group, we associate to each term a numerical weight obtaining by adjusting absolute frequencies by the relative prevalence of the stem in each group. More precisely, let $G_1, \ldots, G_k$ be the groups, and let $c_{i,j} = \sum_{s \in G_i} M_{s,j}$ the absolute frequency of the $j$-th term in the $i$-th group. Defining the global frequency $c_j = \sum_{i=1}^{k} c_{i,j}$, we can determine relative frequencies as $f_{i,j} = \frac{c_{i,j}}{\sum_{r=1}^{p} c_{i,r}}$ and $f_j = \frac{c_j}{\sum_{r=1}^{p} c_r}$. Then, the weight $w_{i,j}$ of each term in each group is given by $w_{i,j} = c_{i,j} \cdot \frac{f_{i,j}}{f_j}$.

Subsequently, we sort the articles within each cluster using an evolutionary algorithm which reminisces a method used for phylogenetic analysis [4]. More, precisely, we subject each cluster to an average-link agglomerative clustering, and we find an arrangement of the resulting tree (that is, determining the left/right disposition of branches at each internal node) in order to minimize the lineal distance between leaves when traversed from left to right. This arrangement is obtained using an elitist generational evolutionary algorithm that evolves bitstrings of $\ell$ bits, where $\ell$ is the number of internal nodes in the tree, each bit indicating the left/right disposition of branches at the corresponding node. The parameters of the algorithm are $popsize= 50$, $maxevals = 25000$, one-point crossover with $p_X = .9$, and bit-flip mutation with $p_m = 1/\ell$.

## 3    Results

Our query results in 374 papers whose thematic base is the application of metaheuristic techniques to multiobjective portfolio optimization. Figure 1 (a) shows a word cloud that illustrates the absolute frequency of stems in $V$. Obviously, stems associated to the search terms stand out, but we can observe the presence of other associated terms that allow determining relevant related topics (e.g., _risk_, _multiobjective_ or _fuzziness_). These terms are also useful to determine the further arrangement of articles into groups. After running the $k$-means algorithm for a different number of clusters, the silhouette metric suggests 5 is the best number of clusters. Figures 1 (b)–(f) show these clusters and the terms they contain. The size of each of these clusters and the most significant terms in each of them are shown in Table 1.

---

[4] https://dblp.uni-trier.de/

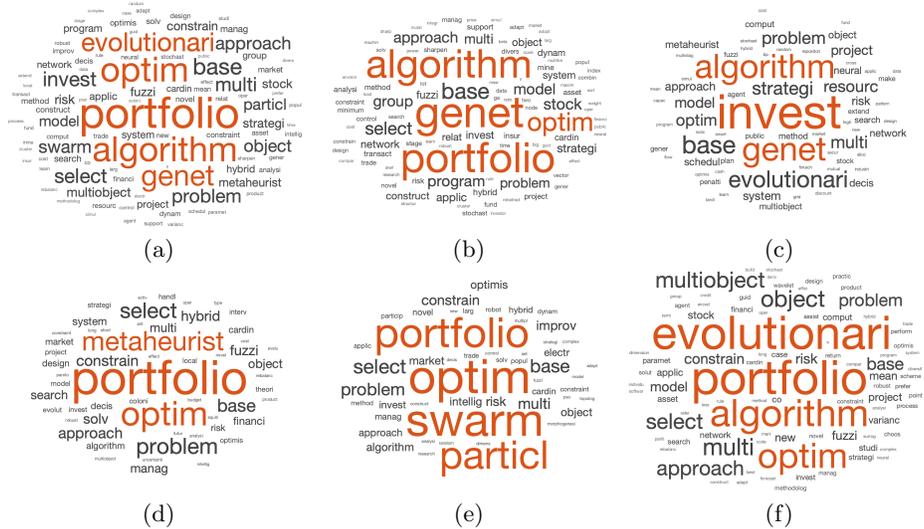

**Fig. 1.** Word clouds for the whole collection of papers (a) and for each of the five groups identified (b)–(f).

**Table 1.** Groups and most significant terms in each of them.

|  | Grupo 1 |  | Grupo 2 |  | Grupo 3 |  | Grupo 4 |  | Grupo 5 |  |
|---|---|---|---|---|---|---|---|---|---|---|
| size | 113 | (30.5%) | 55 | (14.7%) | 49 | (13.1%) | 56 | (15.0%) | 100 | (26.7%) |
| stem (weight) | genet | 698.27 | invest | 419.08 | metaheurist | 162.59 | swarm | 14779.77 | evolutionari | 1120.95 |
|  | algorithm | 148.79 | resourc | 85.37 | portfolio | 38.17 | particl | 4403.45 | multiobject | 137.24 |
|  | group | 144.66 | schedul | 26.35 | evolut | 36.48 | optim | 121.78 | portfolio | 105.12 |
|  | portfolio | 96.20 | genet | 23.12 | solv | 24.83 | electr | 113.73 | varianc | 77.20 |
|  | program | 72.33 | strategi | 18.22 | optim | 18.47 | improv | 44.58 | algorithm | 66.62 |

We observe that the two largest groups (groups 1 and 5) gravitate around genetic programming and multiobjective evolutionary algorithms respectively. This is interesting since our query did not directly include any terms associated with multiobjective optimization, yet the automated analysis was capable of singling out this important approach for portfolio optimization. The three remaining groups are comparatively smaller and are organized around different techniques. For example, group 3 is characterized by the term _metaheuristic_. As a matter of fact, we can find other terms such as _ant_ within the 10 most significant (weight

**Table 2.** Excerpt of article ordering in group 2.

| |
|---|
| Data Transformation Methods for Genetic-Algorithm-Based Investment Decisions. |
| A Decision Investment Model to Design Manufacturing Systems based on a genetic algorithm and Monte-Carlo simulation. |
| A Sustainable Energy Investment Planning Model Based on the Micro-Grid Concept Using Recent Metaheuristic Optimization Algorithms. |
| Maximisation of investment profits - An approach to MACD based on genetic algorithms and fuzzy logic. |
| Extraction of investment strategies based on moving averages – A genetic algorithm approach. |
| Optimizing investment strategies based on companies earnings using genetic algorithms. |
| Genetic Algorithm-based Optimal Investment Scheduling for Public Rental Housing Projects in South Korea. |
| Multi-objective Optimal Public Investment - An Extended Model and Genetic Algorithm-Based Case Study. |

= 16.21). Note that particle swarm optimization seems to follow in group 4, though. Thus, swarm intelligence methods are spread among these two groups. As to group 2, it has a strong genetic algorithm component, but that is not the most distinctive term there. Indeed, the term _neural_ appears in the 10-th position (weight = 13.17), suggesting a larger algorithmic heterogeneity in this group. In fact, this group looks more oriented toward financial issues rather than to algorithmic considerations. Thus, terms related to investment planning and typology, such as _investment_, _resource_ and _scheduling_, stand out. Term _mutual_, corresponding to an actual type of investment fund can be also found in this group (position 19, weight = 12.65).

We have lastly performed an arrangement of articles with each group in order to ease the revision of the bibliography. Table 2 shows an excerpt of this ordering. Obviously, this has been done using a heuristic procedure so there is no optimality guarantees, plus it may be possible to improve the results by adjusting the parameters of the algorithm or using a different underlying metaheuristic altogether.

# Benchmarking Representations of Individuals in Grammar-Guided Genetic Programming ⋆

Leon Ingelse[0000−0001−6067−6318], Guilherme Espada[0000−0001−8128−7397], and Alcides Fonseca[0000−0002−0879−4015]

LASIGE, Faculdade de Ciências da Universidade de Lisboa, Portugal

**Abstract.** Grammar-Guided Genetic Programming (GGGP) has two popular flavors, Context-Free Grammar GP (CFG-GP) and Grammatical Evolution (GE).

In this paper, we first review the advantages and disadvantages of both GE and CFG-GP established in the literature. Then, we identify three new advantages of CFG-GP over GE: direct evaluation, in-node storage, and deduplication. We conclude with the need to further evaluate the comparative performance of CFG-GP and GE.

**Keywords:** Grammar-Guided GP · Derivation Trees · Grammatical Evolution.

## 1 Introduction

Genetic Programming (GP) is praised for its ability to produce useful solutions from vast solution spaces. Being a search-based method, GP performs best when the solution space can be restricted without excluding valid solutions. Grammar-Guided GP (GGGP) [9] uses grammars to solve these issues.

In current research, GGGP has two main approaches for genotype representation. While both approaches use an individual representation that is eventually interpreted as a tree, they differ drastically on the representation. Originally, individuals were represented as derivation trees [9], in a method named Context-Free-Grammar GP (CFG-GP). This approach uses the grammar throughout tree construction, mutation, and/or cross-over operations. The other GGGP approach, called Grammatical Evolution (GE) [7], represents and manipulates individuals as linear strings. Currently, GE is "one of the most widely applied GP methods" [3]. The grammar defined in GE is used to translate these representations to individuals in a process called *genotype-to-phenotype mapping*.

GE has found a strong foothold in GGGP, mainly due to its easier implementation and faster performance of mutation and cross-over operators.

---

⋆ This work was supported by *Fundação para a Ciência e Tecnologia* (FCT) in the LASIGE Research Unit under the ref. (UIDB/00408/2020 and UIDP/00408/2020) and in the FCT PhD scholarship under ref. (UI/BD/151179/2021), by the CMU–Portugal project CAMELOT, (LISBOA-01-0247-FEDER-045915), and the RAP project under the reference (EXPL/CCI-COM/1306/2021).

In this paper we attempt to compare both approaches. First, we present the advantages of GE over CFG-GP in section 2, such as an easier implementation, faster performance of mutation and cross-over operators and being able to reuse work from the broader field of Evolutionary Computation [3]. Then, we highlight the underexposed advantages of CFG-GP, such as better performance [8], the high locality and the predictable effect of mutation and cross-over operations [6], and the guaranteed validity of created individuals in section 3. Finally, we conclude with a discussion in section 4.

## 2    Advantages of Grammatical Evolution

CFG-GP is the most direct approach to encoding grammars, as genotype and phenotype are aligned. However, using different representations for the genotype and the phenotype has advantages. Originally, the introduction of GE as a replacement of CFG-GP was mainly motivated by individuals being smaller [7], which reduces memory usage and allows GGGP to be more scalable. Furthermore, there are advantages of GE, due to the *mechanical sympathy* of computers in regard to linear strings. Mainly, GE avoids pointer chasing, making mutation and cross-over operators more efficient. Moreover, the generation of an individual, which effectively consists of generating a random array, can be done faster than the CPU can send the bits to memory.

Correspondingly, on the algorithm implementation side, mutation and cross-over are more easily implemented: Mutation entails the selecting of a random location in the string and updating the value of that location. For cross-over, a location is randomly selected, and two individuals are cut and then concatenated at that location, producing two individuals.

The separation of genotype and phenotype allows the user to "decouple the search engine from the problem [at hand]", so that the same algorithm can be easily applied to different domains [2]. As such, GE also allows the user to reuse research from the areas of Genetic Algorithms and Evolution Strategies [3].

Later, the same authors found populations in GE to be more diverse, because genotypical differences do not necessarily translate to phenotypical differences in GE [4]. As genotypes may contain a lot of redundant information, multiple genotypes can translate to the same phenotype. This phenomenon is called high redundancy. However, this high redundancy is a disadvantage: 90% of mutations do not have any effect on the phenotype, rendering the mutation useless [6].

In the same study, they showed that the mutations that did have effect, often produced child individuals very dissimilar to their parents, resulting in *low locality*. This is because changing the value in one location of a linear string, can cascade to the production of other parts of the individual (fig. 1). High redundancy and low locality result in the fitness of GE resembling a Random Search algorithm [6].

There have been attempts to diminish above-mentioned disadvantages. To make sure genotypical differences affect the phenotype, redundant parts of an individual can be trimmed [7]. By trimming, a maximum length is set for each linear string. If a genotype, after being operated on, needs more genes than its length allows for, one can apply the *wrapper* approach and continue translation from the start of the linear string [4]. Problematically, it results in a lower locality.

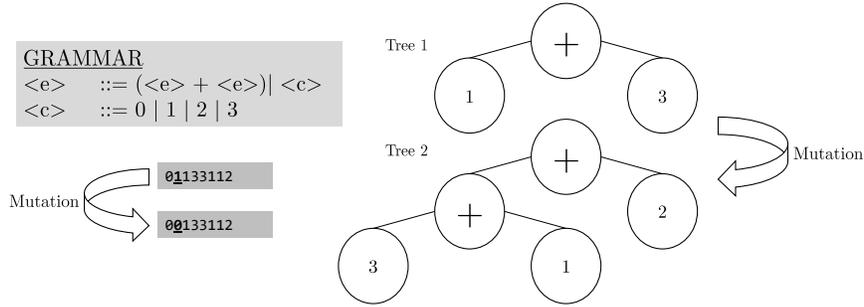

**Fig. 1.** Example where a mutation cascades to another part of the individual. The second number is mutated, referring to the left node with value 1 in tree 1. This mutation cascades to the right node as well.

Furthermore, the wrapper approach might result to invalid-individual generation. Look at the grammar fig. 1 and consider the genotype 000. This genotype maps to never-ending plus operations. The fitness of this individual cannot be calculated. Methods such as assigning invalid individuals a low fitness, repairing them, and phenotype-validity checking, are costly and break with the simplicity of GE. Initial populations often show 70% invalid individuals [5].

To improve on the locality of evolved populations, Structured Grammatical Evolution (SGE) was proposed [1]. In SGE, individuals are represented by lists of linear representations. Each list contains all information of one production in the grammar. As such, mutation only affects that single production, and minimizes the cascading to other parts of the individual. A comparison with normal GE shows SGE performing better [1, 2]. Note that SGE breaks with the simplicity of GE, diminishing certain GE advantages, such as a simple implementation. Moreover, individual representations are more complex and take up more memory as every individual requires the space of the largest possible one.

## 3   Advantages of using Derivation Trees as the Genotype

In section 2, we have presented the three main issues of GE, non-effective mutations, low-locality, and non-valid individuals. CFG-GP does not suffer from these issues as the genotype and the phenotype are aligned. These advantages are put forward to argue that CFG-GP performs better than GE [8, 2].

Besides the direct advantages resulting from the genotype-phenotype alignment, we identify three less discussed advantages of derivation trees: direct evaluation, in-node storage, and deduplication.

**Direct Evaluation:** During fitness evaluation in GE, each individual is first translated to a string (the program), which is then parsed, and finally evaluated.

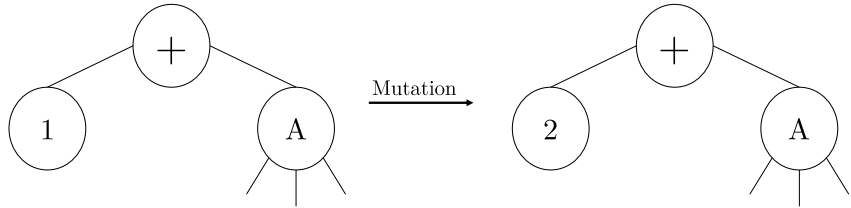

**Fig. 2.** Example where in-node storage of evaluation is beneficial. The value 1 is mutated to 2. The mutated tree can be easily reevaluated by summing 2 and the evaluation of subtree A.

Crucially, translation and parsing are not free, both in execution time and memory consumption. In languages with a just-in-time compiler, this cost is exacerbated even further, as it has extra work to do tracking freshly generated code. In CFG-GP, derivation trees are traversed only once, evaluating the program directly, with no parsing required, saving time and memory.

**In-Node Storage:** Since trees consist of nodes in memory, meta-information can be stored in each node. One useful example would be to store the partial evaluation of that node, to reuse in trees that share genomic material with the current one. Caching evaluation results (and other meta-information) is beneficial to a large class of problems, including symbolic regression. Further evaluations can avoid re-evaluating subtrees (c.f. fig. 2).

**Deduplication:** Because of cross-over, different individuals frequently have subtrees in common, that do not need to occupy duplicated space in memory. Applying deduplication also has the side effect of improving caching (if used in combination with in-node storage) when a given subtree occurs in different individuals. Thus, each subtree needs only be evaluated once, even if it is used in different individuals. This is incredibly difficult in the GE approach: a tree node might not even correspond to an executable element by itself.

## 4   Discussion

Practitioners should be aware of these advantages when selecting which approach to take. In fact, these decisions are relevant when designing a GGGP framework, and not when implementing a domain-specific grammar. Because one of the main advantages of GE is the ease of implementation of crossover and mutation operators, we argue that a single investment in implementing the CGP-GP operators can be worthwhile as it can be applied to different domains. The second advantage of CGP-GP is the memory saved on the individual representation, but we identify that this might not compensate the extra memory necessary for parsing or the potential time savings by caching evaluation or memory saved by deduplicating common subtrees. Furthermore, despite being less widely used, CFG-GP can perform better than GE [8, 6].

Future research should consider a wide-range comparison of both approaches, covering usability, performance, maintainability, interpretation and scalability.

# Implementing an Evolutionary Algorithm to Optimise Fractal Patterns and Investigate its Possible Contribution to the Design of Engineering Systems[*]

Habiba Akter[0000−0002−6873−7549], Rupert Young[0000−0002−1669−2393],
Phil Birch[0000−0002−7740−9379], and Chris Chatwin[0000−0001−9371−8502]

Department of Engineering and Design,
School of Engineering and Informatics,
University of Sussex,
United Kingdom
h.akter@sussex.ac.uk
r.c.d.young@sussex.ac.uk
p.m.birch@sussex.ac.uk
c.r.chatwin@sussex.ac.uk

**Abstract.** This is ongoing interdisciplinary research drawing inspiration from the rapidly growing field of evolutionary developmental biology, i.e., the so-called "Evo-Devo" paradigm. We not only aim to investigate the similarities but importantly, also the differences between the synthesis of a biological organism and an engineering system. A bespoke algorithm will be developed using an Evolutionary Algorithm to generate fractal patterns observed in nature. The success of this method will allow us to investigate its possible usage in designing an engineering system.

**Keywords:** Evolutionary Algorithm · Fractals · Engineering Design.

## 1 Introduction

The study of Evolutionary Algorithms (EAs), in particular micro-evolutionary processes, has broadened the area of research into the evolution of the phenotypic structures of biological organisms [1]. We aim to implement an EA to generate commonly observed patterns observed in nature. Although classical geometry tries to describe these patterns, it is often adequate to specify the more complex shapes [2]. Michael Barnsley's work has studied and implemented Iterated Function Systems (IFS) to generate such complex phenotypic structures which are commonly observed in biological organisms, particularly plants[3,4]. Since both IFS and EA are iterative in nature, we link them together to evolve fractal patterns. Engineering is also similar in some ways, particularly from a design perspective. For example, it would be immensely hard to create a modern bicycle from scratch. The design has 'evolved' over the last two hundred years and has become progressively refined (and specialised e.g. racing bikes versus mountain bikes etc). We aim to implement EA and fractal analysis to design an engineering system as well.

---

[*] EVO* 2022 - Proceedings in ArXiv - Late-Breaking Abstracts

## 2   Bespoke Algorithm

The algorithm we propose uses two main components:

- Evolutionary Algorithm (EA)
- Iterated Function System (IFS)

The pseudo algorithm 1 gives an describes the overall structure.

---

**Algorithm 1** EA-IFS for Fractals

---

**PARAMETERS:**
*Current iteration, itr*
*Iteration limit, $i_{max}$*
*Iteration limit for IFS, $it_{max}$*
**OUTPUT:** Fractal Phenotype, *img*

1:  $P \leftarrow$ initial population
2:  $N \leftarrow$ size of $P$
3:  $P' \leftarrow P$ with maximum $F$
4:  $P_f \leftarrow$ final set of Population
5:  **while** $itr \leq i_{max}$ **do**
6:      Initialise $P_n$
7:      Evaluate and sort $P$
8:      **function** Crossover($P$)
9:          $C \leftarrow$ Offspring after Crossover
10:     **end function**
11:     **function** Mutation($P$)
12:         $M \leftarrow$ Offspring after Mutation
13:     **end function**
14:     $P_f \leftarrow P' \cup C \cup M$
15:     Evaluate $P_f$ and find $P'$
16:     Plot $img$ using $P'$ in $IFS$
17:     $itr \leftarrow (itr + 1)$
18: **end while**
        **return** $img$

---

The algorithm is run for a pre-selected number of iterations. The initial population set, $P$, generated by the EA is an array of matrices representing the coefficients of an IFS. The population goes through the steps of evaluation, reproduction (crossover and mutation) and selection, finally generating the fittest chromosome, $P'$. Then $P'$ is passed to the IFS to generate the final phenotype. Thorough tests and evaluations will confirm the final parameters to be employed for the EA. These parameters include the size of $P$, crossover and mutation probability, crossover and mutation rate, number of iterations for the IFS and the terminating condition of the algorithm. It is noteworthy that the parameter values for each phenotype may vary. Also, different selection mechanisms need to be investigated.

## 3    Progress

We have designed and evaluated an algorithm to produce Barnsley fern fractal pattern. Out of several different fractal dimension mechanisms, [5,6,7], we chose to use the box-counting dimension to automatically assess the fitness of the phenotype in a physically meaningful manner. Figure 1 shows some initial attempts at generating the fern.

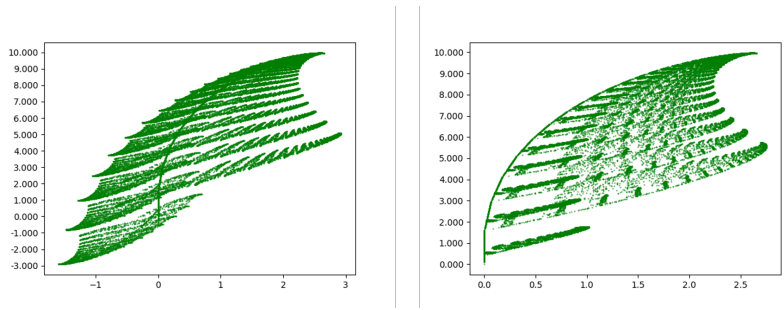

**Fig. 1: Initial attempts of generating Barnsley fern**

With time, the algorithm has been improved to produce more natural fern representation as shown in Figure 2.

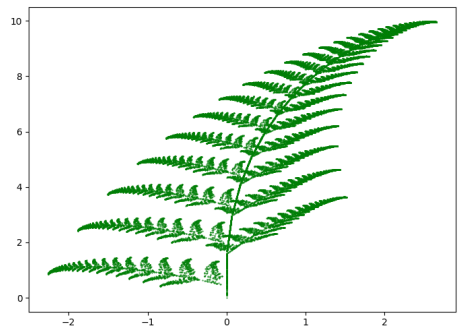

**Fig. 2: Fern generated by the improved algorithm**

## 4    Conclusion and Future work

The initial attempts of the proposed algorithm have successfully produced the fractal structure of the Barnsley fern. The box-counting dimension is employed as the fitness cost to allow reach an optimal solution generated by the Evolutionary Algorithm. Building on this success, the next step will involve extending the modelling of the phenotype to generate fully three-dimensional structures. This will allow structures of engineering interest to be investigated in detail.

### Acknowledgement

This work was funded by the Leverhulme Trust Research Project Grant RPG-2019-269 which the authors gratefully acknowledge.

# Statistical Investigation of Neighbourhood Utility in a Parallel Machine Scheduling Problem

André L. Maravilha[1][0000−0002−5869−3052], Letícia Mayra-Pereira[2], and
Felipe Campelo[3][0000−0001−8432−4325]

[1] Centro Federal de Educação Tecnológica de Minas Gerais
Divinópolis 35503-822, MG Brazil. [`andrelms@ufmg.br`]
[2] Graduate Program in Electrical Engineering, Universidade Federal de Minas Gerais
Belo Horizonte 31270-901, MG Brazil. [`leticiamayra@ufmg.br`]
[3] College of Engineering and Physical Sciences, Aston University
Birmingham B4 7ET, UK. [`f.campelo@aston.ac.uk`]

**Abstract.** Local search heuristics are commonly used for several classes of combinatorial problems, and neighbourhood functions play a critical role in their ability to adequately explore the search space. We suggest a statistical approach to the investigation of neighbourhood utility for specific algorithm-problem pairs. The approach is illustrated using a Simulated Annealing with six neighbourhood structures for the Unrelated Parallel Machine Scheduling Problem with Sequence Dependent Setup Times. The proposed approach may be useful for devising self-adaptation rules for the selection of neighbourhood functions conditionally on problem characteristics and the search stage.

**Keywords:** Combinatorial Optimisation· Landscape Analysis· Statistical Modelling

## 1 Introduction

The unrelated parallel machine scheduling problem with machine and sequence-dependent setup times, commonly represented as $R|s_{ijk}|C_{max}$ [2,4], is an NP-hard problem [1] informally defined as follows: Given a set of jobs $J = \{1, 2, \ldots, n\}$ and a set of machines $M = \{1, 2, \ldots, m\}$, each job $i \in J$ has to be processed by a single machine $k \in M$. Preemption is not allowed, i.e., once a machine starts to process a job it cannot be interrupted. In addition, let $p_{i,k} > 0$ be the time required by machine $k \in M$ to process job $i \in J$; and $s_{i,j,k}$ be the setup time required to prepare machine $k \in M$ if jobs $(i, j) \in J$ are both processed at $k$, with job $j$ immediately after job $i$. The objective in this problem is to assign all jobs $j \in J$ to machines $k \in M$, sequencing the jobs in each machine so as to minimize the completion time of the last job to leave the system (i.e., the makespan $C_{max}$). A formal definition of the problem is provided in [7].

Given an incumbent solution, it is known that the makespan can be improved only if the makespan machine[1] is changed, either by altering the sequence of jobs processed at the makespan machine, or through the reassignment of a job currently assigned to the makespan machine to another machine. Changes that do not involve the makespan machine are not able to improve the makespan of a solution, and may be "neutral" (not change the makespan) or "deleterious" (increase the makespan). The number of distinct solutions presenting the same makespan value is usually large, and the objective function landscape is commonly characterized by neighbourhoods in which most changes in the incumbent solution result in no immediate impact on the makespan. This flatness in the fitness landscape tends to degrade the performance of local search-based heuristics. Metaheuristics commonly employ diversification mechanisms [3, 5] or secondary solution quality criteria (e.g. the sum of completion times of all machines) [6] to reduce this problem.

---

[1] The machine executing the last job to be finished.

In this work we investigate six neighbourhood functions commonly employed in local search heuristics for the $R|s_{ijk}|C_{max}$ problem: Shift, Switch, Task Move, Swap, Direct Swap and Two-Shift [6]. As mentioned earlier, the proportion of solutions that result in makespan improvement for this problem class is generally very low for all neighbourhood structures. To investigate how this can affect the performance of metaheuristics used for the solution of this particular problem class, we perform an experimental investigation on how the expected utility of a single move using each of the six neighbourhood structures mentioned above changes, as a function of problem size (number of machines / jobs) and of the point at which the movement is performed. The results of this investigation can then be used to propose statistical strategies for adapting the probabilities of using each neighbourhood at different points of the search.

## 2 Experimental Protocol

In this experiment, the six neighbourhood functions mentioned in the previous section are explored at different stages of the search for several instances of the problem, using a bespoke adaptation of Santos *et al.*'s Simulated Annealing [6].[2] The algorithm was run on the 200 training instances described by Vallada and Ruiz [7], with dimensions given by all combinations of $M \in \{10, 15, 20, 25, 30\}$ machines, $J \in \{50, 100, 150, 200, 250\}$ jobs, and maximum setup time $S \in \{9, 49, 99, 124\}$. The setup times of each instance were generated using a uniform distribution between one and $S$, as described in the original reference [7]. For each problem size $< M, J, S >$, two distinct instances were available, and the algorithm was run once on each instance.

At each run, any time the incumbent solution changed, a complete enumeration was performed for each neighbourhood function. For all neighbours generated using each neighbourhood function, we collected: (i) the elapsed time and iteration number; (ii) $C_{max}$ of the incumbent solution; (iii) Sum of processing times (SPT) of the incumbent solution, $s_x$; (iv) Neighbourhood size (cardinality) and count of neighbours that present improvements in $C_{max}$; and (v) Best, mean and worst changes in $C_{max}$. Changes in terms of quality values were measured as the percent improvement of the neighbour over the incumbent, $\delta_\%(x') = [f(x) - f(x')]/f(x)$, with $f(x)$ and $f(x')$ being the makespan of the incumbent and neighbour solutions, respectively. This *improvement ratio* represents the proportion of solutions in a given neighbourhood that result in improvements over the incumbent one, which can be interpreted as the probability of observing an improvement after a single movement using a given neighbourhood function. If $\eta_i(x)$ denotes the set of neighbours of incumbent solution $x$ according to the $i$-th neighbourhood function and $\eta_i^*(x) \subset \eta_i(x)$ is the subset of neighbours that improve on the incumbent, then $\pi_i = |\eta_i^*(x)| / |\eta_i(x)|$ can be interpreted as the probability of a randomly sampled neighbour $x' \in \eta_i(x)$ resulting in an improvement over $x$. For the subset of neighbours that result in improvement, the expected magnitude of the improvement measures how much the makespan is improved:

$$E\left[\delta_\%(x')|x' \in \eta_i(x)\right] = \frac{1}{|\eta_i^*|} \sum_{k \in \eta_i^*} \left[\frac{f(x) - f(x')}{f(x)}\right]. \tag{1}$$

---

[2] Full source code is available at https://github.com/andremaravilha/upmsp

Given the above, the *expected utility* of a given neighbourhood can be seen as the expected improvement after a single perturbation according to that neighbourhood function,

$$E\left[u_i\right] = \pi_i \times E\left[\delta_\%(x')|x' \in \eta_i(x)\right] = \frac{1}{|\eta_i|} \sum_{k \in \eta_i^*} \left[\frac{f(x) - f(x')}{f(x)}\right]. \qquad (2)$$

Using the data gathered using the experimental protocol detailed earlier in this section, the expected utility was calculated for each neighbourhood function every time the incumbent solution changed. This allowed us to model the effects of problem size (number of jobs and machines), as well as the effect of (normalized) time and other covariates on the utility of each neighbourhood, enabling the analysis and modelling described in the next section.

## 3  Preliminary results

Preliminary results, shown in Figure 3, indicate a clear decrease of $E\left[u_i\right]$ with time for all neighbourhood functions, which is expected since it becomes more difficult to improve over incumbent solutions as the search progresses. Neither the initial utilities nor the rates of decrease, however, were homogeneous: certain neighbourhoods, such as Shift and Switch, tend to maintain higher utilities throughout the search; while others, such as Swap and Two-Shift, have their expected utilities drop strongly in the latter portions of the optimisation process, across all problem sizes. Other interesting dependencies on $M$ and $J$ are also suggested in this exploratory visualisation. To quantify these effects we fit a set of regression models to the data, aiming at obtaining a statistical model capable of estimating the expected utility of each neighbourhood based on known problem features and the state of the search, so as to bias the choice of $\eta_i$ towards higher-payoff moves. After initial model exploration, the following model form was chosen and fit independently for each neighbourhood:

$$\log_{10}\left(E\left[u_i\right]\right) = \beta_0 + \beta_M M + \beta_J J + \beta_S S + \beta_t t' + \beta_{s_x} s'_x + \beta_{(tt)} t'^2 + \beta_{(MM)} M^2 + \beta_{(JJ)} J^2 + \beta_{(SS)} S^2 +$$
$$\beta_{(s_x s_x)} s'^2_x + \beta_{(Jt)} Jt' + \beta_{(JM)} JM + \beta_{(JS)} JS + \beta_{(MS)} MS + \beta_{(Ms_x)} Ms'_x + \beta_{(MS)} MS \qquad (3)$$

where $t' = \log_{10} t$ ($t$ is the normalised time) and $s'_x = \log_{10} s_x$ is the logarithm of the sum of processing times for the incumbent solution.

Based on these models, we propose a simple modification to the neighbourhood selection procedure in Santos *et al.*'s Simulated Annealing, in which the expected utility values of the movements are used to bias the choice of movement to be executed at any iteration. The standard version of the algorithm selects the movements probabilistically, with each movement having a fixed probability of $p_i = 1/6$. Instead, we propose calculating these probabilities as:

$$p_i = \frac{1 - p_{max} + w_i\left(|H|p_{max} - 1\right)}{|H| - 1}, \qquad (4)$$

where $H = \left\{\eta_1, \ldots, \eta_{|H|}\right\}$ is the set of available neighbourhood functions, $p_{max}$ is an upper limit on the selection probability of any individual neighbourhood, and

$$w_i = \frac{E[u_i]}{\sum_{j=1}^{|H|} E[u_j]}$$

with $E[u_i]$ calculated using the regression model (3).

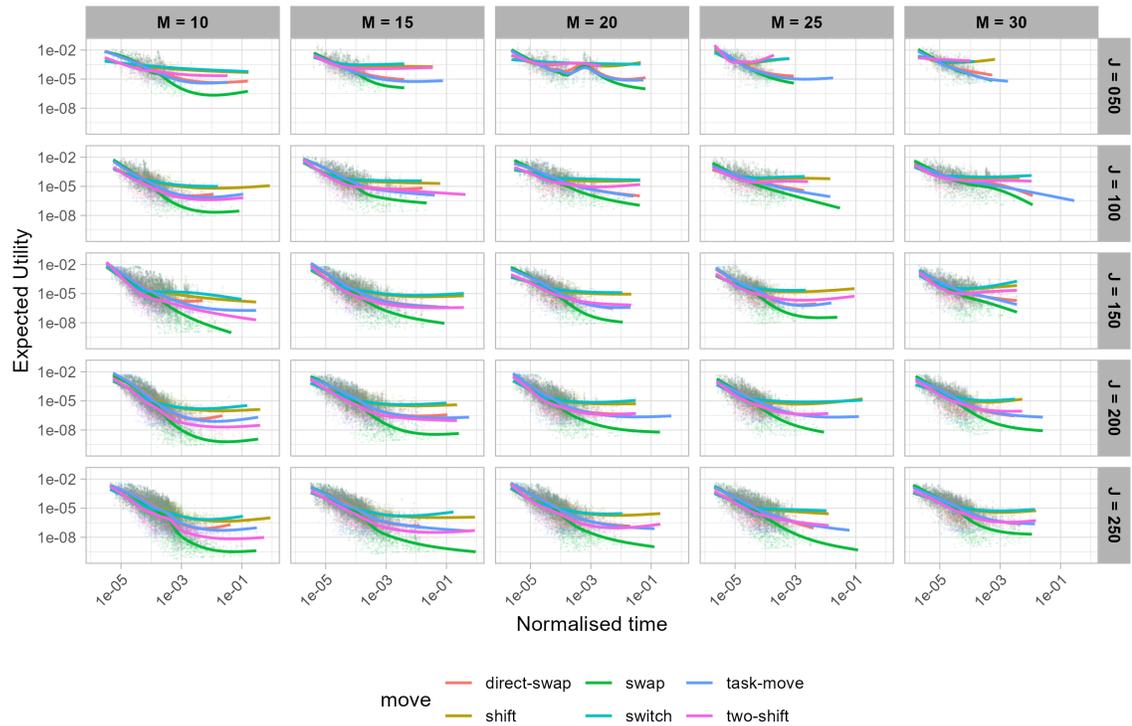

**Fig. 1.** Expected utility value of executing a movement according to each neighbourhood function, conditioned on number of machines $M$, number of jobs $J$ and (normalised) time.

# Using Evolutionary Algorithms for Multi-Constrained Path Calculation in a Tunnelled Network Topology*

Habiba Akter[1][0000−0002−6873−7549], Chris Phillips[2]

1 Department of Engineering and Design,
School of Engineering and Informatics,
University of Sussex,
East Sussex, United Kingdom
h.akter@sussex.ac.uk
2 School of Electronic Engineering and Computer Science
Queen Mary University of London
London, United Kingdom
chris.i.phillips@qmul.ac.uk

**Abstract.** Bio-inspired computation is implemented in a wide range of research fields. We propose the use of Evolutionary Algorithms (EA) to develop a path computation tool to send data over the network in the presence of Internet tunnels. We have chosen this algorithm for its evident success as a search and optimisation tool.

**Keywords:** Evolutionary Algorithm · Multi-Objective Evolutionary Algorithm · Multi-Objective Optimisation · Path Computation.

## 1 Introduction

Since the 1970s, a number of evolutionary methods have been suggested, including Genetic Algorithm (GA), Evolutionary Strategy (ES) and Evolutionary Programming (EP) [1] and their applications in science and engineering have been growing rapidly. The success of Evolutionary Algorithms (EAs) in finding solutions to problems with multiple objectives has been the main motivation for us to use them in developing a path calculation tool for an internet topology where the paths have multiple constraints. The main novelty in this is the implementation of an EA path computation for the presence of internet tunnels, which can be implemented using different technologies [2]. The use of a Genetic Algorithm(GA) in finding the shortest path has been proved to be possible in the works [3,4,5]. A Multi-Objective Evolutionary Algorithm (MOEA) can also be implemented to make the tool more efficient [6].

## 2 Proposed Tool

The tool we propose generates a network topology having optional tunnels implemented in the domains. Then we use an EA to calculate paths from a user-selected source node, $S$ to a destination node, $D$ to send data over. Each path in the topology has two constraints associated with it:

- Average end-to-end delay
- Financial cost

---

* EVO* 2022 - Proceedings in ArXiv - Late-Breaking Abstracts

Hence, the problem of the path calculation becomes multi-objective where the objectives will be to:

1. Minimise the delay and
2. Minimise the cost

Thus, the problem becomes a "minimisation" problem. We will investigate the efficiency of EAs to find an optimal path both by keeping it as a Multi-Objective Optimisation Problem (MOOP) and by converting it to a Single Objective Optimisation Problem.

The pseudo algorithm 1 describes the overall structure.

---

**Algorithm 1** Path Computation Tool

---

**PARAMETERS:**
*Current iteration, itr*
*Iteration limit, $i_{max}$*
*Iteration limit for IFS, $it_{max}$*
**INPUT:**
*Source and Destination, $S, D$* **OUTPUT:** Optimal path, $P'$

1: $G \leftarrow$ network topology
2: $P \leftarrow$ initial population
3: $P' \leftarrow P$ with minimum $F$
4: $P_s \leftarrow Secondary population$
5: $P_f \leftarrow$ final set of Population
6: Generate $G$
7: **while** $itr \leq i_{max}$ **do**
8:      Initialise $P_n$
9:      Evaluate and sort $P$
10:     **function** CROSSOVER($P$)
11:         $C \leftarrow$ Offspring after Crossover
12:     **end function**
13:     **function** MUTATION($P$)
14:         $M \leftarrow$ Offspring after Mutation
15:     **end function**
16:     Generate $P_s$
17:     $P_f \leftarrow P' \cup C \cup M \cup P_s$
18:     Evaluate $P_f$ and find $P'$
19:     $itr \leftarrow (itr + 1)$
20: **end while**
        **return** $P'$

---

The number of iterations for the EA in the tool is pre-selected. The initial population set, $P$, holds all possible paths between a source node and destination node, $S$ and $D$. These paths then undergo evaluation, reproduction (crossover and mutation) and selection. A number of random paths are inserted as a secondary set of populations. All these are merged making the final population set, $P_f$ which is then passed to the next iteration as its initial population. At each iteration, $P_f$ needs to be evaluated and shortest path $P'$ is obtained which has the minimum score. The $P'$ of the final iteration is the output shortest path.

## 3   Work Progress

We have already proved the significance of tunnels being present in a part of the network topology where the usage of tunnels involves financial cost [7,8]. Not only that, we have also successfully designed an initial tool that implements GA. Figure 1 shows a sample network topology generated by the tool.

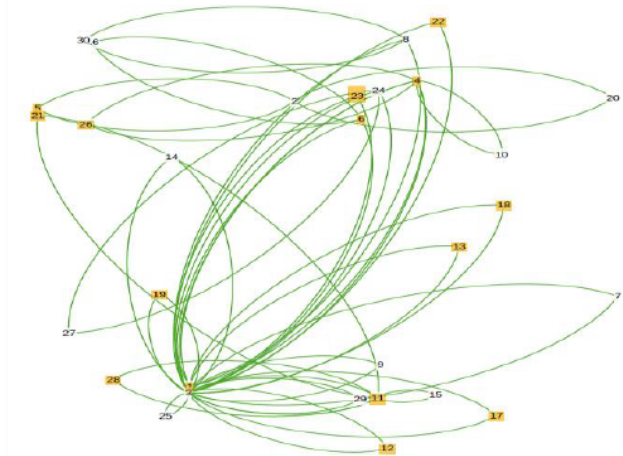

**Fig. 1: An example topology of 30 nodes**

We have run the GA for 50 iterations that generates a set of best possible paths to send data from node 12 to node 16. To evaluate the tool, we have first programmed it to calculate all best possible paths. The tool outputs the paths as follows:

– $12 \rightarrow 2 \rightarrow 22 \rightarrow 3 \rightarrow 17 \rightarrow 4 \rightarrow 27 \rightarrow 5 \rightarrow 16$
– $12 \rightarrow 3 \rightarrow 26 \rightarrow 4 \rightarrow 7 \rightarrow 2 \rightarrow 5 \rightarrow 16$
– $12 \rightarrow 2 \rightarrow 5 \rightarrow 16$
– $12 \rightarrow 18 \rightarrow 3 \rightarrow 30 \rightarrow 7 \rightarrow 2 \rightarrow 5 \rightarrow 16$

   – $12 \rightarrow 2 \rightarrow 22 \rightarrow 3 \rightarrow 10 \rightarrow 4 \rightarrow 27 \rightarrow 5 \rightarrow 16$

It is then improved to output the shortest path, $12 \rightarrow 2 \rightarrow 5 \rightarrow 16$

## 4   Conclusion and Future work

The initial attempts of the proposed tool have been tested and evaluated. We have designed fitness equations to reach an optimal solution generated by the Genetic Algorithm. Based on this success, the next step will involve the implementation of a Multi-Objective Evolutionary Algorithm which will include designing new equations for fitness functions.

# Evolving Artificial Spin Ice Geometries Towards Computing Specific Functions

Arthur Penty and Gunnar Tufte

`arthur.penty@ntnu.no`

Norwegian University of Science and Technology (NTNU)

Artificial Spin Ice (ASI), meaning a collection of interacting nanomagnets, were originally constructed as a model system to study physical phenomena, but have become a promising avenue for *in materio* computation. Though the potential of ASI for computation is often cited within the ASI community, there are few concrete examples of it being used as such. Here we demonstrate evolving the geometry of an ASI such that it can perform a simple classification task. The success of the method on this simple problem gives confidence that this methodology could be used to create ASI capable of solving more complex tasks.

## 1  Artificial Spin Ice

ASI [9] refers to a collection of interacting nanomagnets, so named as its origin was as an 'artificial' model system for studying the structure of water molecules in ice. Magnets in an ASI are small and elongated causing them to be bi-stable, thus the spin state of a magnet (magnetisation direction) can be represented as a 0 or 1. We can refer to the ensemble of these individual spin values as the state of the ASI. The spin of a magnet can flip to the other state given sufficient pressure from its neighbours, or by the use of an external magnetic field. The state of an ASI is an emergent property. Through tuning the positions and orientations of the nanomagnets, interesting large scale patterns can be observed arising from low-level interactions of the magnets. We refer to a particular layout of the magnets in an ASI as an ASI geometry.

Recently ASI has become of interest in it own right. Material scientist are interested in studying ASI as a meta-material, while computer scientists see potential in harnessing its extremely low-energy state-transitions as a mechanism for computation. The majority of ASI research focuses on a small hand-full of different ASI geometries, with simple repeated patterns. Research on the computation aspects of ASI tends to take one of these well-studied geometries and evaluate them using computational frameworks such as Reservoir Computing (RC) [2,3]. Alternatively, nanomagnetic logic research [5], explores hand-crafted, carefully engineered geometries which implement basic logical units such as logic gates and counters. In contrast to these approaches, our interest is in exploring novel complex ASI geometries for computation, exploiting the emergence of complex dynamics, similar to RC, beyond the constraints given by repeated patterns, whilst moving away from the top-down approach of nanomagnetic logic to a bottom-up evolutionary search.

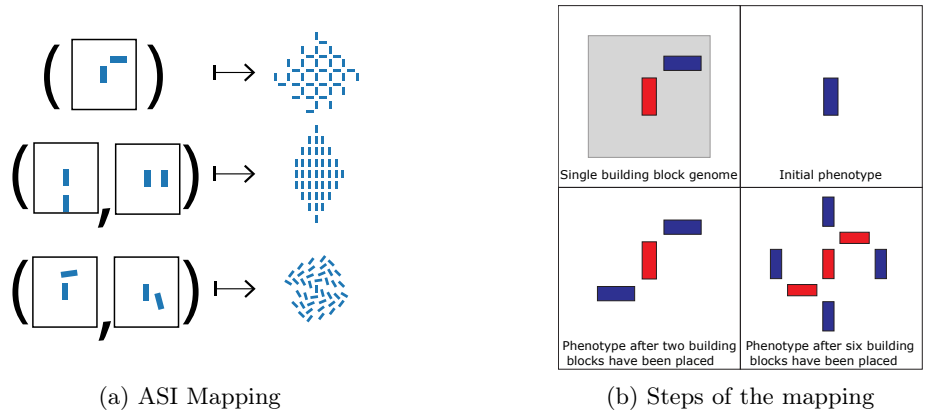

(a) ASI Mapping                  (b) Steps of the mapping

Fig. 1: (a) Building block genotypes and their corresponding phenotypes. (b) Stages of the genotype to phenotype mapping of the first genome shown in (a)

## 2 Evolving geometry

Evolutionary methods have long been used to explore novelty and diversity in form. Karl Sims's Virtual Creatures [7], Richard Dawkin's Biomorphs [1] and even the breadth of different 'designs' apparent in the natural world illustrate the ability of evolution to find new, well-suited and interesting forms. Previously, we have detailed a method for representing and evolving ASI geometries [6]. For the sake of brevity, we give only a rough outline of the representation here.

At the heart of any Evolutionary Algorithm (EA) is a population which, though variation and selection, converges towards some desirable behaviour. Individuals in our population represent ASI geometries. Specifically, our individuals are an ordered collections of building blocks from which ASI geometries are constructed through a generative process. The building blocks, as shown in Fig. 1a, consist of exactly two magnets. ASI are constructed from these building block through a deterministic genotype to phenotype mapping. Starting from a single magnet the geometries are grown by adding one building block at a time. A building block is rotated and positioned such that one of its magnets perfectly overlaps a magnet already in the geometry. Fig. 1b illustrates the development steps for a simple geometry, and a more rigorous description of the process and representation can be found in [6]. The geometries can be mutated through mutating their constituent building blocks and the resulting phenotypes can be evaluated using an ASI simulator such as flatspin[4].

## 3 Searching for computation

As an example of a computational task, we have selected the problem of classification. In classification, the goal is to sort a set of $n$ inputs into $k$ different bins based on some features of the input. A classifier must be able to discriminate on some aspects of the input but disregard other aspects (unless $n = k$). We take a very simple, but non-trivial instance of a classification problem: given a set of 4-bit binary numbers, classify the numbers by their value modulo 4. Our case is a special case of classification as the input space is finite and small enough that all possible inputs can be tested exhaustively.

To map this abstract problem onto our physical system, we encode the 4-bit number by perturbing an ASI with a series of magnetic field applications. For each bit in the input we apply a field with its strength given by a single period of a sine wave. If the bit is 0 the field is applied at $0°$, otherwise it is applied at $90°$. After all 4 field applications, the final state of the ASI is recorded. This is repeated for each possible 4-bit input. In the perfect case, the final states of the ASI under each input should be identical to the final state of the other inputs in its class, but not identical to any others. We construct our fitness function by penalising each deviation from the perfect solution.

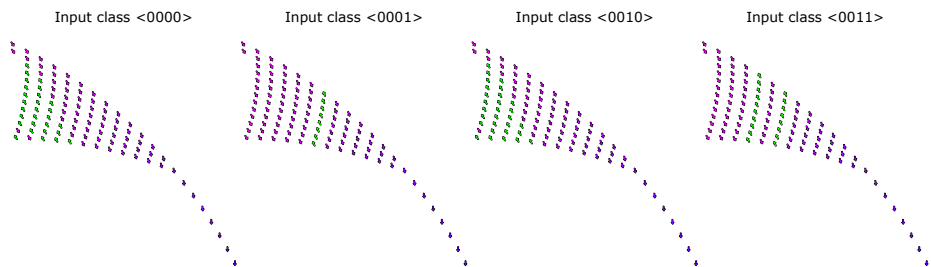

Fig. 2: Final state of the perfect solution. Each magnet is the ASI is represented as an arrow, with colour and direction representing the magnet's spin. Only one representative from each class is shown, as they are identical within the same class.

A simple EA of population size 100 was run using the representation and fitness function described and we constrain our geometries to contain exactly 100 magnets. The EA terminated at generation 254 due to the discovery of a geometry with perfect fitness. Fig. 2 shows the geometry that achieved perfect fitness in its four different final states.

## 4 Discussion

We have demonstrated that, using our representation of a ASI geometry, it is possible to find an ASI which perform a specified function though evolving its geometry. Despite this being very simple classification task, it shows the potential of ASI as a classifier, as well as the effectiveness of evolving geometries as a means to imbue ASI with a desired computational function.

Looking at Fig. 2, we can see the final state for the input classes <0000> and <0010> are extremely similar, differing by only one magnet. This is likely an artefact of evolution taking the path of least resistance, and in simulation having these two states be so close is not an issue for classification. However one could envision, in a physical realisation of such a classifier, having greater difference between the states could make reading out the state from the system easier and more reliable. As such, it would be an interesting extension of this work to modify the fitness function that it encouraged the final states of each class of inputs to have greater differences.

**Acknowledgements:** This work was funded in part by the Norwegian Research Council IKTPLUSS project SOCRATES (Grant no. 270961), and in part by the EU FET-Open RIA project SpinENGINE (Grant no. 861618). Simulations were executed in flatspin[4] on the NTNU EPIC compute cluster[8].

# On the difficulties for evolving 4-part harmony

F. Fernández de Vega
fcofdez@unex.es

University of Extremadura, Spain

**Abstract.** In this paper we present an improvement over previous attempts for addressing evolutionary 4-part harmony. We describe first how the set of rules used in the fitness function, required in this kind of composition technique, has been enlarged, thus allowing augmented sixth chords, Neapolitan chord, as well as ninth chords. This greatly increases the search space. We analyse the difficulties for addressing the problem, describe some attempts to improve quality of results and finally show an example evolved using the new approach.

**Keywords:** First keyword · Second keyword · Another keyword.

## 1 Introduction

4-part harmony is a well known composition area that every music student must face when pursuing a professional or higher music degree.

Although the problem has been already addressed more or less successfully, only recently a software tool that includes all the standard harmony rules has been published, under the supervision of professional music teachers. This project, Sharpmony [1], provides information about all the standard harmony rules that are applied when checking if a given exercise is correct or not. As we may notice, 20 different categories are shown describing the rules applied, such as *parallel 5ths*, and some of the categories includes a number of exceptions and cases that every student must master when finishing their studies. For instance, for the specific case of *Wrong note duplication*, whose error color assigned by Sharpmony is Fuchsia, three different cases are considered, and some of them with two additional exceptions. So, near 50 different controls must be checked for every couple of chords when analyzing a score.

To the best of our knowledge, in the latest attempt to address this problem by evolutionary means [1], less than half of the above described controls where implemented as part of the fitness function: only 11 rules were applied. Yet, for the problem addressed, no perfect solution was found, and depending on the approaches tested, a fitness value among 10 to 40, which corresponds to the number of errors, were found after several hours of the evolutionary process trying to devise a good solution.

---

[1] https://sharpmony.unex.es

## 2   The problems in the problem

### 2.1   Available solutions

Of course, when we add restrictions over a previous version of a given problem, the number of solutions greatly decreases. And this is also the case for the problem we address.

If we focus on the example shown in above referred paper, as shown in figure 1, we may check one of the solution that was found then, that allowed the use of secondary dominants together with other simpler techniques that allowed to reach the solution found in figure 2. But checking that solution using the new set of rules included in the fitness function we see that the number of errors displayed in figure 2, as notes with colors assigned, is larger that 50.

So, what was a reasonable solution with the reduce set of rules, is not a solution anymore with the enlarged set of rules, and this makes the search process much harder.

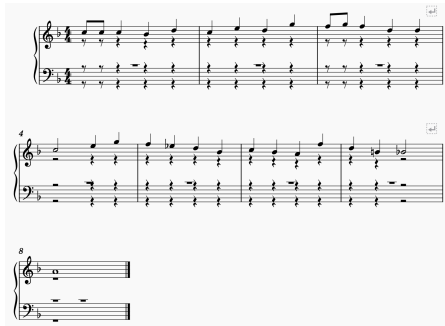

**Fig. 1.** Melody to be harmonized.

### 2.2   The search space

On the other hand, when the search space grows, the possibilities for finding a good solutions reduces. In the paper referred above, secondary dominants were included in the set of available chords, together with the standards I,II...VII degrees corresponding chords. But now, we have also included chords with diatonic seventh, ninth chords (minor and major), augmented sixth chords and the Neapolitan chord. Thus, the number of available chords is more than double when compared with the previous approaches, thus notably increasing the size of the search spaces.

Therefore the challenge is double: on the one hand we see that given the larger number of rules, the number of available solutions decreases; secondly, the search space significantly grows, and thus the time required for finding solutions will grow. But maybe this could be somehow alleviated with the new kind of chords available: they might provide new solutions that were not available before.

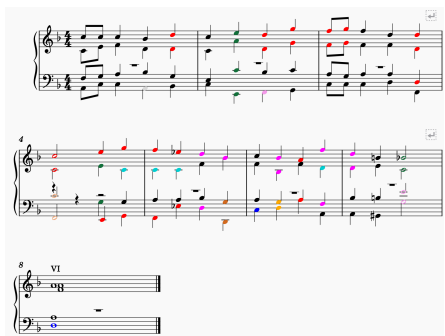

**Fig. 2.** Solution evolved using 2017 approaches: more than 50 errors detected using 2022 updated fitness function. Check sharpmony.unex.es for color codes

This paper presents preliminary results considering the expanded possibilities described above. Although modulation processes are already allowed, in this first series of experiments we kept them out of the road map.

## 3   The new approach

As described above, the number of available possibilities is large:

- I,II,III,IV,V7, VI, VII degrees in a given key: Given that 4 different notes are considered per chord (first, third, fifth and seventh), and the rules to shape proper chords: the fundamental must be included always, but the other ones may or may not repeat, and for every choice different positions may be assigned to each note: more than 200 possibilities are available per possible chord.
- In the case of V7 we may add the ninth suppressing any of the already included notes (any but the fundamental).
- From each of the chords described above, notes can be assigned to any of the octaves available 2,3, 4 or 5. Thus for each chord, and considering the range voices may feature, we number of possibilities per chord notes configuration are between 20 and 30 once we assign octaves to notes.
- Augmented sixth chords and Neapolitan chord are new possibilities that were not available before.

Thus, a huge number of possibilities per chord are available, and if we want the find good chord progressions, only considering that 2 chords are analyzed, and several thousand possibilities per chord can be used, millions of progressions are available in the search space.

In order to narrow the search process, two different techniques are applied here: On the one hand i) The progression of chords to be applied to every note is evolved first, thus reducing the size of the search space. ii) A local-search process is applied to reduce the number of errors that the random selection of notes may introduce when a genetic operation is applied.

## 4   Experiments & Results

We run the evolutionary process for the same problem shown above, and after 10 generations -about 4 hours-, using 4 individuals, and only mutation+selection, we find a solution with only 6 errors. Figure 3 shows the solution found. We are confident that this new approach can be further improved to include modulation process in the future.

## 5   Acknowledgments

This publication is part of the Grant PID2020-115570GB-C21 funded by MCIN/AEI/10.13039/501100011033, and Grant GR21108 funded by Junta de Extremadura and the European Regional Development Fund, a way to build Europe.

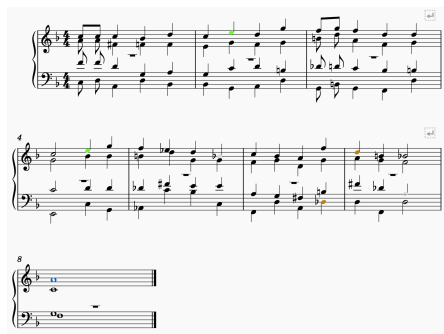

**Fig. 3.** Solution evolved and checked using 2022 updated fitness function.

# Explaining Protein-Protein Interaction Predictions with Genetic Programming

Rita T. Sousa, Sara Silva, and Catia Pesquita

LASIGE, Faculdade de Ciências da Universidade de Lisboa
{risousa,sgsilva,clpesquita}@ciencias.ulisboa.pt

**Abstract.** Explainability is crucial to support the adoption of machine learning as a tool for scientific discovery. In the biomedical domain, ontologies and knowledge graphs are a unique opportunity to explore domain knowledge, but most knowledge graph-based approaches employ graph embeddings, which are not explainable. However, when the prediction target is finding a relation between two entities represented in the graph, such as in the case of protein-protein interaction prediction, semantic similarity presents itself as a natural explanatory mechanism. This work uses genetic programming over a set of semantic similarity values, each describing a semantic aspect represented in the knowledge graph, to generate global and interpretable explanations for protein-protein interaction prediction. Our experiments reveal that genetic programming algorithms coupled with semantic similarity produce global models relevant to understanding the biological phenomena.

**Keywords:** Explainable Artificial Intelligence· Knowledge Graph · Genetic Programming · Protein-Protein Interaction Prediction.

## 1   Introduction

In artificial intelligence (AI) applications in science, explanations are crucial not only for the user's trust but also for discovering of new knowledge. Several explainable artificial intelligence (XAI) approaches have been proposed, but only a few approaches integrate domain knowledge modelled through semantic technologies such as ontologies and knowledge graphs (KGs) [1]. However, most KG-based machine learning (ML) approaches apply KG embedding methods which are sub-symbolic representations of KG entities that are not interpretable by default [2].

Since similarity assessment is a natural explanatory mechanism [8], an alternative explanatory strategy is to use the ontologies and KGs to measure the semantic similarity (SS) between entities in the graph. This is particularly relevant in the biomedical domain where ontologies allow the description of complex biological phenomena, providing the scaffolding for comparing biological entities through their ontology representations. Furthermore, since SS can be computed using different portions of the KG to reflect different semantic aspects (SA) [5], we propose that SS can provide more granular explanations with higher information content.

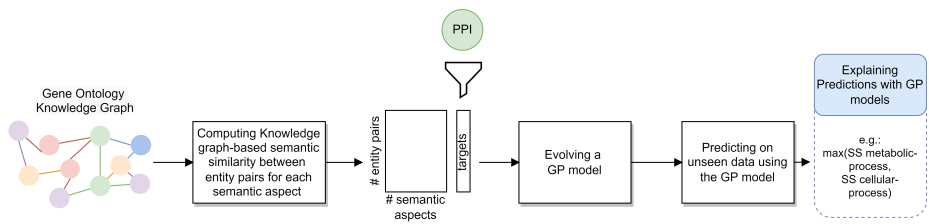

**Fig. 1.** Overview of our methodology.

We address the explainability problem for protein-protein interaction (PPI) prediction by using genetic programming (GP) algorithms [3] over ontology-based SS values that capture different SAs. GP algorithms were chosen given their ability to produce potentially interpretable models that provide a global explanation of how the model works, unlike many classical ML algorithms and deep learning methods.

## 2    Methods

PPI prediction is cast as a classification task that takes as input a KG and a PPI dataset containing a set of protein pairs that interact (Figure 1). The PPI dataset was obtained from the STRING Database[1]. We used the Gene Ontology (GO) [7] KG composed by the ontology and annotations that link proteins to GO functions. The functions in GO are described concerning three domains: biological processes, molecular functions and cellular components. The three GO domains are represented as root classes. The child classes of these roots, such as *'cellular_process'* and *'catalytic_activity'*, represent an ontology SA.

We followed the methodology of [6]. First we computed SS scores for the protein pairs according to different ontology SAs corresponding to the 50 subgraphs rooted in the direct subclasses of GO roots (removing aspects with potential bias for PPI, namely *'binding'* and *'protein-containing_complex'*). We used the $ResnikMax_{Seco}$ similarity measure implemented in [6]. Then, we evolved a GP model to predict PPI for protein pairs represented by their 50 SS scores.

Although GP searches the space using genetic operators that manipulate their syntactic representation fulfilling every constraint for transparency, sometimes the solutions grow exponentially with each generation, and the interpretability is lost. To tackle this, we modified the fitness function of standard GP to penalize solutions with a depth greater than six (value given as a parameter), thus lowering the probability of deep trees. In addition, this variation of GP (GP6x), only uses interpretable operators, namely maximum, minimum, addition and subtraction. Operators such as multiplication and division were excluded as it is difficult to interpret the biological meaning of multiplication/division between SS values. We performed 10-fold cross-validation and compared GP (no depth penalization and 6 operators) with GP6x.

---

[1] https://string-db.org

**Table 1.** Example of an explainable GP6x model and description of protein pairs SS with supporting evidence for interaction status (dark blue: molecular function, light blue: cellular component, pink: biological process).

---

**Explainable Model:**

$\max(SS_{\text{multicellular\_organismal\_process}}, SS_{\text{cellular\_process}}, SS_{\text{molecular\_adaptor\_activity}}, SS_{\text{signaling}},$
$SS_{\text{molecular\_function\_regulator}}, SS_{\text{catalytic\_activity}}, SS_{\text{behavior}} + SS_{\text{immune\_system\_process}})$

---

**40S ribosomal protein S12 − 40S ribosomal protein S10** $(+/+)$

40S ribosomal protein S12 and 40S ribosomal protein S10 are components of the 40S ribosomal subunit that plays a central role in protein translation and is characterized by multiple binding sites.

---

**S100-A10 protein − neuroblast differentiation-associated protein** $(+/-)$

Protein S100-A10 works together with neuroblast differentiation-associated protein AHNAK in the development of the intracellular membrane.

---

**Kinetochore-associated protein 1 − Tubulin beta-6 chain** $(-/-)$

Kinetochore-associated protein 1 and tubulin beta-6 chain are both located in the nucleus but have different functions: while the first one is envolved in mitosis, the late is envolved in GTP binding.

---

**Protransforming growth factor $\alpha$ − Disks large homolog 2** $(-/+)$

TGF-$\alpha$ is a mitogenic polypeptide, and disks large homolog 2 is a member of the membrane-associated guanylate kinase. Both participate in MAPK cascade.

---

## 3 Results

The median weighted average of F-measures (WAF) for GP is 0.875 while for GP6x it is 0.866 ($p$-value of 0.0065 with Kruskal-Wallis test). As to the number of nodes in the unsimplified models, the medians are 49 for GP and 17 for GP6x ($p$-value of 0.0004). With small differences, all GP6x models consider maximum similarities of multiple SAs with a majority describing biological processes. This corroborates prior knowledge that for two proteins to interact they usually participate in the same biological processes.

To investigate the trade-off between performance and explainability we chose four protein pairs and analysed the input SS values and one of the GP6x models and its predictions, in Table 1. The protein pairs were randomly chosen among well-predicted positive pairs $(+/+)$, well-predicted negative pairs $(-/-)$, wrong-predicted positive pairs $(+/-)$, and wrong-predicted negative pairs $(-/+)$. One of the most interesting results is the analysis of the two pairs for which GP6x fails. *S100-A10* protein and the *neuroblast differentiation-associated* protein are known to interact, and the biological processes where both proteins participate are described in the literature. However, according to GO annotations, proteins have the same location in the cell but do not share biological processes, resulting in low SS values for relevant SAs. The misclassification can then be justified by the incomplete annotation of these proteins. Concerning the pair *TGF-α - Disks large homolog 2*, GP6x predicts an interaction given the high similarity values for SAs relevant, but it appears as not interacting in the dataset. It is important to note that the negative PPI dataset examples were generated by negative random sampling. Interestingly, the literature describes interactions between proteins of the same family of the pair, which probably means that this pair is actually positive but not yet present in STRING.

## 4    Conclusion

A significant direction of XAI research is the discussion of trade-offs involving performance prediction and interpretability [4]. Our results show that the performance of the more interpretable methods is lower, but what they sacrifice in performance is gained in explainability. The analysis of the selected examples highlights how explainability can be key to uncover issues with the underlying data and even pose new hypothesis. One of the main advantages of transparent methods, such as GP, is that the explanation is the model itself, avoiding the need for local explanations or post-hoc techniques.

***Acknowledgements*** This work was funded by FCT through LASIGE Research Unit (UIDB/00408/2020, UIDP/00408/2020); projects GADgET (DSAIPA/DS/0022/2018) and BINDER (PTDC/CCI-INF/29168/2017); PhD grant SFRH/BD/145377/2019.

# Towards Brain Controlled Sound Sampling

David DeFilippo

University of California San Diego, San Diego CA 92093, USA
`ddefilip@ucsd.edu`

## 1  Introduction

The musical instrument proposed here infers imagined music from readings of brain activity via machine learning. During training one EEG (electroencephalogram) cap is connected to a Variational Autoencoder (VAE) and a library of sound samples are connected to a second VAE. Time series data from the Brain and the music recorded synchronously are trained on the two separate VAEs. When the instrument is in use, a musician hears, or possibly imagines a sound, and the EEG is rendered in real-time to sample sounds from the latent music space. Mental activity is mapped directly to perceptions, delimited by the latent space of generalized sounds. The plan will dive into the Variational Autoencoder (VAE) and Regressive Flows (RF) as a mapping strategy, where RF creates an invertible map between two trained latent spaces of VAEs. In this case, one VAE contains a latent space of EEG data and the other contains sound samples.

The latent space of the sound samples plays a similar role to biological memory of the sounds heard. The latent space of the EEG, plays a similar role to the mind itself, being the potential sensations. Extended Mind Theory [2] claims that under certain conditions an external device is not something like another mind. Rather it is something that is part of the user's mind. The claim relies in part on the Parity Principle. The Parity Principle suggests that if the external device is sufficiently similar in consequence, so that what is done with it, stands in place of what is done in the head, the use of the device is functionally isomorphic to the mental process. The mind is then extended into the device under functional isomorphism.

The Brain Computer Interface (BCI) device here involves computational recognition of mental activity. What is generated in the mind as a sound, could be considered an expectation about what the device will render. Once the sound is rendered the role of imagination is offloaded to the external actuation of the sound, that being a latent representation of the mental activity of previous imagination. Performed as an instrument a musician generates mental activity that is mapped to its potential auditory sensation. The mental activity names the auditory sensation bypassing the physical movement necessary to produce the sound. A recent study [12], uses EEG combined with Music Information Retrieval techniques to create a data set of both perceived music and imagined music as brain state read outs. The author claims it will soon be possible to recognize a song as a thought in the head.

Functional equivalence between mental imagery and visual perception has been demonstrated since the 1980's by [5]. Though the effect on visual information mechanisms in the brain is smaller than when presented with visual objects, there are similarities that exist on many levels of visual processing. Studies since have found cognitive similarities between mental imagery and visual perception. Another such study [1] finds the two, perceived and imagined objects, rely on the bottom-up processes in the brain that generate mental states based on encoded perceptual information. In other words, the mind is in a state of active inference that generates a mental prediction about the states of the world. More recently, neuroimaging studies have recently been conducted with music. There is a general cognitive equivalence between mental imagery and musical imagery. The studies of musical imagery [6] [13] [8], have found evidence of neural similarities between imagining musical sounds and perceiving musical sounds. Both states of cognition involve recruiting of substrates comprising auditory and motor cortical regions. These studies attempt to isolate the common experience of hearing a tune in your head or a mental concert. On a broader level these studies move toward understanding the internal subjective experience of being conscious in the world.

### 1.1   Brainwaves in the History of Experimental Electronic Music

In the early 1960's, composer Alvin Lucier got an idea from research physicist Edmond Dewan at the Air Force Research Labs. Dewan's subjects were able to consciously alter the intensity of Alpha rhythms in their brain to render morse code, transmitted to a teleprinter. This would lead to *Brainwave music and the chicken doctor: Music for Solo Performer* (1964-65) by Lucier. Here amplified brainwaves were projected through loudspeakers in contact with percussion instruments, causing the instruments to vibrate. When David Tudor was preparing to perform the piece at the University of California Davis in 1967, the two needed a special amplifier and assistance with where to place the electrodes. The doctor from the veterinary school who helped them with the electrodes had only placed electrodes on chickens for experiments, hence the name [7].

## 2   BCI Instrumentation with Two VAEs

The VAE's efficiency as a generalizer means that unrecorded brain states produce unrecorded sounds, for one. However, even when sampling the latent music space with a known the sensation, the VAE also will produce events not in the training set. A VAE is a generative model that captures the underlying probability distribution of a set of data $s$ as p(s), by considering a lower dimensional latent representation of the data $\eta$. The generative model takes the form of $p(s, \eta) = p(s|\eta)p(\eta)$. The VAE instrument described here involves training latent spaces of two VAE's, $\eta^m$ and $\eta^E$, from data $s^{music}$ and $s^{EEG}$.

Modeling the complex distributions of $s$ as real world data with a VAE involves the use of Variational Inference (VI). VI models $s$ as an optimization problem by positing a set of generic densities as normal distributions, $q_\phi(\eta|s)$. This term inverts the likelihood to reveal the statistical relation of hidden causes, $\eta$, to effects, $s$. The symbol $\phi$ denotes that the normal distributions are transformed under the optimization procedure of minimizing the KL-Divergence ($D_{KL}$). $\phi$ encapulates two transformable parameters: the mean $\mu$ and the variance $\Sigma$, where $q_\phi^*(\eta|s) = \underset{q}{argmin} D_{KL}[q_\phi(\eta|s)||p(\eta|s)]$.

The terms are rearranged as in equations 1 and 2 in the ELBO formulation for modeling $s^{EEG}$ and $s^{music}$. The reference material for these equations is found in [4].

$$L_{\theta,\phi} = E[log\, p_\theta(s^{EEG}|\eta^E)] - D_{KL}[q_\phi(\eta^E|s^{EEG})||p_\theta(\eta^E)] \qquad (1)$$

$$L_{\theta,\phi} = E[log\, p_\theta(s^{music}|\eta^m)] - D_{KL}[q_\phi(\eta^s|s^{music})||p_\theta(\eta^m)] \qquad (2)$$

The ELBO, $L_{\theta,\phi}$, depicts model evidence minus reconstruction error. The p() and q() are parameterized by $\theta$ and $\phi$ and subject to transformation. Minimizing the reconstruction error and regularizing $q_\phi(\eta|s)$ is accomplished with the ELBO formulation. The evidence from $s$ is $log\, p(s)$. To model the evidence of $\eta$, $log\, p(s)$ is expressed as $E[log p_\theta(s|\eta)]$, to express the likelihood of obtaining an $s$ given an $\eta$. Subtracting the reconstruction error from the model evidence is said to minimize the reconstruction error. While the divergence between $q_\phi(\eta|s)$ and $p_\theta(z)$ regularizes $q_\phi(\eta|s)$.

The encoder $Q$ uses $q_\phi(\eta|s)$ to encode $s$ into a latent representation $\eta$ and the decoder P uses $p_\theta(s|\eta)$ to generate a $s$ given an $\eta$. A set of data, $s$, is encoded by an encoder $Q$ for its mean, $\mu$ and variance $\Sigma$. The means and variances make up a series of gaussian distributions with $\mu$ as the center in its latent space and $\Sigma$ representing the spread of the distribution around that center. The network's ability to generalize from $s$ is attributed to the inclusion of the variance measure to represent the training data. A new technique for better organization of a latent space is termed a Normalizing Flow. The technique creates a smooth invertible mapping between latent dimensions (of a single latent space) and transforms the KL objective $q_\phi(\eta|s)$ also known as the approximate posterior into a more nuanced multi-modal density, which is shown to better emulate the complexity of data sourced from the real world [9]. In the next section, the notion of generating $s$ with an $\eta$ from a different latent space, by way of an invertible mapping between latent spaces is described. Here two VAE's with two latent spaces $\eta^m$ and $\eta^E$, such that $s^{EEG}$ generates sound from $\eta^m$, not an EEG signal from $\eta^E$.

## 2.1  An Invertible Mapping Technique Between Two Latent Spaces

The Normalizing Flow is similar to another technique called a Regression Flow [4], which creates an invertible map between two different latent spaces. Here the normalizing and regression flow are described with respect to the problem at hand, mapping latent EEG to latent music. In [4], a sample of audio generates parameters to control a synthesizer, described as parameter inference. This tool named the Flow Synthesizer, functions within Digital Audio Workstation wrapping the functionality of VST synthesizer plugins. With a graphical user interface, users can supply the model with any kind of sound and obtain a parameter setting that produces a reasonable approximation of that sound.

Here with brain controlled sound sampling, the inverse is considered such that, given parameters as an EEG signal, the audio itself is generated. The auditory events are sounded as inference from the states of the mind, acting as the control parameters. This could be termed auditory inference. To produce auditory inference, we consider the two latent spaces, $\eta^m$ as the auditory space and $\eta^E$ as the EEG space. By the invertible mapping technique, given an $\eta^E$ we generate an s not from the corresponding EEG space, but from the auditory space, $\eta^m$. Without inscribing flows into the operation, the joint likelihood has the form in equation 3.

$$log\, p_\theta(s, \eta^m, \eta^E) = log(p_\theta(s|\eta^m, \eta^E)p_\theta(v)) + log\, p_\theta(\eta^m|\eta^E) \qquad (3)$$

With this formulation the crucial inference problem of $log\, p_\theta(\eta^m|\eta^E)$ can be rendered independently from the variational approximation by optimization. For the inference of $log\, p_\theta(\eta^m|\eta^E)$ a transformation, $f_\psi$, represents the main flow component. Optimization locates the parameter $\psi$ so that the latent $\eta^m = f_\psi(\eta^E)) + \epsilon$. The term $\epsilon$ models the inference error, where $\epsilon$ is approximated by N(0,$C_e$) which is a zero mean gaussian with covariance, $C_e$. Decomposing $C_e$ gives another hyperparameter, $\lambda$ necessary to optimize in the full joint likelihood of equation 4.

$$L_{f_\psi, \lambda} = log[p_\theta(\eta^m|f_\psi, \lambda, \eta^E)p_\theta(f_\psi|\eta^E)p_\theta(\lambda|\eta^E)] \qquad (4)$$

With two posteriors to optimize $p_\theta(f_\psi|\eta^E)$ and $p_\theta(\lambda|\eta^E)$ the resolution rests on the use of VI. Thus an approximate posterior for a KL objective is defined as $q_\phi(f_\psi|\eta^m, \eta^E)$. Factoring the objective leaves the final inference problem, an ELBO being equivalent to the evidence plus two divergences to reach the optimization targets.

$$L_{f_\psi, \lambda} = log[p_\theta(\eta^m|f_\psi, \lambda, \eta^E)] + D_{KL}[q_\phi(f_\psi|\eta^m, \eta^E)||p_\theta(f_\psi|\eta^E)] + ...$$
$$D_{KL}[q_\phi(\lambda|\eta^m, \eta^E)||p_\theta(\lambda|\eta^E)] \qquad (5)$$

The KL objective involving the optimization of $\lambda$ in equation 6 can be resolved by implementing gaussian distributions for both the prior and posterior. These distributions account for the mean and the variance of each. Again, $\lambda$ is necessary for modeling the inference error, $\epsilon$, associated with the transform, $f$, of the latent $\eta^m$, in the crucial inference $log\, p_\theta(\eta^m|\eta^E)$. With the other KL objective involving $\psi$, the main parameter of the transform, $f$, such that $\eta^m = f_\psi(\eta^E) + \epsilon$, can not be assumed as a gaussian because it maps between the two latent spaces and requires a highly non-linear, multi-modal transfer function, so the inference, $log\, p_\theta(\eta^m|\eta^e)$ is smooth. To do this the posterior, $q_\phi(f_\psi|\eta^m, \eta^E)$, is parameterized with a flow $k$, $q_k(\eta_k^m)$. The procedure transforms the posterior via a succession of composed maps where the number of those maps in succession is $k$. See figure 2 for a general example. A distribution is given by this procedure, computing the determinant of a Jacobi matrix, which contains the partial derivatives of $f$ and the latent $\eta^E$. The first divergence in equation 5 then becomes:

$$D_{KL}[q_\phi(f_\psi|\eta^m)||p_\theta(f_\psi|\eta^E)] =$$

$$E_{q_0}[logq_0(\eta_0^m)] - E_{q_0}[log\,p(\eta_k^m)] - E_{q_0}\left[\sum_{i=1}^{k}log\left|det\,\frac{\partial f_i}{\partial\eta^E_{i-1}}\right|\right] \quad (6)$$

The quantity $\eta^m$ now is a transformed version of $\eta^E$ and the invertible map is in place. With that, the extent possible of the technical plan for the BCI instrument is described. For future research, a real-time implementation could be investigated. With two latent spaces and an invertible map, the encoder for one VAE could theoretically be used to compress a real world phenomena to the size of a lower dimensional sample to decode a point from the second latent space. This is different than current methodology with one VAE that involves applying a random sample to the low dimension of the latent space to decode a compressed representation of the state.

# Study on Genetic Algorithm Approaches to Improve an Autonomous Agent for a Fighting Game*

N. Escalera[1], A.M. Mora[1], P. García-Sánchez[2]

[1] Departamento de Teoría de la Señal, Telemática y Comunicaciones.
ETSIIT-CITIC, Universidad de Granada.
`escaleranm@gmail.com,amorag@ugr.es`
[2] Depto. Arquitectura y Tecnología de Computadores.
ETSIIT-CITIC, Universidad de Granada.
`pablogarcia@ugr.es`

## 1 Introduction

A fighting game is a 1 vs 1 confrontation in which the players (characters) try to reduce the health bar of the opponent by punching or kicking him. Players can normally also do special movements, which are more complicated to execute, but which can reduce the rival's health in a higher amount. Artificial Intelligence (AI) engines in this type of games have a big handicap in comparison to other genres, since they must yield very fast decisions, given the highly dynamic rival movements.

In this paper, we have created a Non-Player Character (or bot) aiming to beat to any opponent (being a human player or another NPC). To this end, we have started from a competitive bot (from the state of the art) having as AI engine a set of rules, depending on some conditions, threshold and weights. Then we have optimized these values by means of different schemes of Genetic Algorithms (GAs) [1].

Fighting games have been a very prolific research area, in which many different approaches have been proposed: script-based bots [2,3], Monte-Carlo Tree Search [4] and, recently, Deep Reinforcement Learning ones [5,6]. Evolutionary Algorithms have been also applied in this domain [7,8], however, to our knowledge, just a very few number of studies, such as [9], have focused on Evolutionary approaches for AI.

Thus, this work presents a preliminary study on the improvements in behaviour and performance obtained through the application of GAs over an AI engine for an agent/bot for fighting videogames. It can be considered as a first step in our research on this domain, given our previous and successful experiences in other games [10–12].

* This work has been partially funded by projects PID2020-113462RB-I00 (ANIMALI-COS), granted by Ministerio Español de Economía y Competitividad; projects P18-RT-4830 and A-TIC-608-UGR20 granted by Junta de Andalucía, and project B-TIC-402-UGR18 (FEDER and Junta de Andalucía)

The obtained agents (one per GA variation) have been tested against a set of rivals, some of the prefabricated and some created by other authors of previous editions of the competition.

We have used a simulator called FightingICE[3], which is the one used in the international Fighting Game AI Competition (FGAIC). FightingICE also offers a framework where researchers can develop and test their own agents. It also includes some prefabricated bots that can be used to fight against (and test) our own agents (see Figure 1).

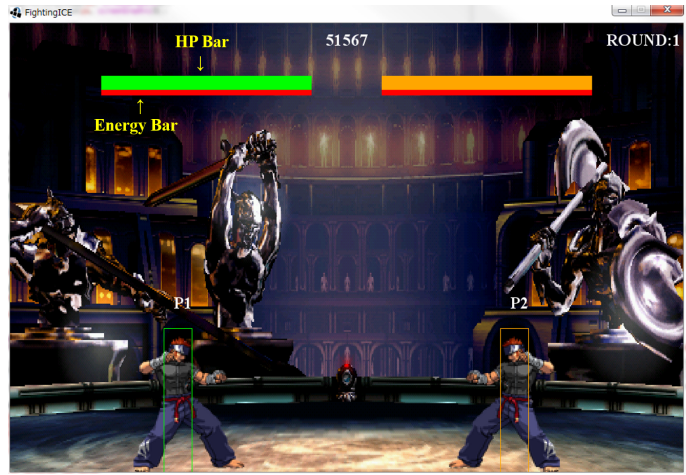

**Fig. 1.** Example snapshot of FightingICE

## 2   Implemented agent

For this study we have considered the standard health/time mode in which we have to reduce the opponent's health bar to 0 or to reduce it more than what it can do to our bot in the limited amount of time per combat. The inputs for the bots could be three buttons (Punch, Kick and Special) or one of the 9 directions in a control pad or joystick. We have considered the same character for all the agents, named Zen.

We have implemented a modified version of Mizuno's bot [13], named *MizunoAI*. It was an entry submitted to 2014 FGAIC, based on a fuzzy controller together with a classification method (kNN) to select the best action to perform after some simulations which alternates between kNN-based decisions and fuzzy system-based behaviour based on a random value. The optimization considers 10 parameters such as distances and health levels to make decisions on the actions to perform.

This genetic bot has been implemented in Java and interacts with Fighting-ICE framework through different external files. The GA optimizes the set of parameters which the AI engine depends on. Thus, every individual is a vector of 10 values that 'models' a behaviour for the agent (depending on the values it has). Every individual is evaluated in a set of 6 game combats against two possible rivals. After these combats, a fitness value is calculated as the average of $remaining\_time * (health_{agent} - health_{rival})$ of the performed matches.

Two different GA approaches have been implemented:

*Generational without elitism*: in which the whole population could be replaced. Binary tournament is run considering all the individuals in the population (1 vs 1) and the best in their respective combats will remain. Uniform crossover and random mutation of one gene (with a probability) are again executed. The population will be the generated offspring + half random individuals. Thus, we aim for a high exploration factor with this approach.

*Generational with elitism*: We have added elitism to this approach, so, instead of generating half of population random, all the individuals combat in pairs, surviving the winners to make the crossover. In addition, the best individual of every generation will always survive.

## 3   Experiments and Results

The configuration considered in the experiments is: 20 generations, 16 individuals in population, crossover probability = 60%, mutation probability 10%, number of combats for fitness = 6, players' health = 300. Two different rivals have been used in the fitness computation: Dora (an advanced agent from the competition) and BCP (very aggressive bot also from the competition). We have made several runs with the different GA schemes and using different rivals in the fitness computation. After this optimization phase, we have chosen the best individual in each experiment confronting the best of every run in a tournament against the rest, the winner of 9 combats will be finally selected.

Then, the chosen agents have fought against other bots. In Table 1 it can be seen a summary of the percentage of wins for each one obtained in a series of 9 combats. The percentage of victories represented is the one obtained by the agent in the row against the one in the column.

As it can be seen, the winning rates are not very surprising, given that the original Mizuno is able to win a majority of combats against the optimized versions. The reason could strive in the fact that the optimization uses as rival BCP and Dora, which are quite different and less competent in behaviour to Mizuno. Thus, the obtained (optimized) agents would not work properly against this though agent. In the same line, the results of the evolved bots against other agents are neither good, so they are not able to win against them (but in just a few cases).

Although we have not managed to obtain remarkable victory rates, Table 2 illustrates there is some result improvement with respect to the base agent. The results show that, although the optimization has not been enough to win a large number of fights, it has allowed us to improve practically all the results, with *MizunoAIEV-DoraBCP-Elitism* being the configuration with the best results. It is interesting to note that the fact that *BCP* yields such negative life differences is related to its behavior, i.e., it is an agent that once it achieves its objective, which is to lock up the opponent, it is very easy to win, so in its victories it wins by a large difference. We also highlight the fact that we have usually obtained better results from the elitist configurations than from the non-elitist ones, this is due to the fact that in the non-elitist configurations there is a very high variability and given the noisy nature of the problem, it is likely to yield false good solutions.

|  | MizunoAIOrig | BCP | Dora | Thunder |
|---|---|---|---|---|
| MizunoAIOrig | - | 0% | 0% | 0% |
| MizunoAIEV-DoraBCP-Elitism | 0% | 22.22% | 0% | 0% |
| MizunoAIEV-DoraBCP-NoElitism | 0% | 11.11% | 0% | 0% |
| MizunoAIEV-BCP-Elitism | 0% | 0% | 0% | 0% |
| MizunoAIEV-BCP-NoElitism | 33.33% | 0% | 0% | 0% |
| MizunoAIEV-Dora-Elitism | 33.33% | 0% | 0% | 0% |
| MizunoAIEV-Dora-NoElitism | 33.33% | 0% | 0% | 0% |

**Table 1.** Percentage of agent wins against state-of-the-art bots. The names in rows belong to our optimized (EV) agents. They indicate the rival in the fitness computation, and the GA scheme used.

|  | BCP | Dora | Thunder |
|---|---|---|---|
| MizunoAIOrig | -134.44 | -285.67 | -279.67 |
| MizunoAIEV-DoraBCP-Elitism | -103.77 | -269.22 | -210.56 |
| MizunoAIEV-DoraBCP-NoElitism | -167.67 | -232.56 | -218.89 |
| MizunoAIEV-BCP-Elitism | -146.33 | -212.33 | -230.33 |
| MizunoAIEV-BCP-NoElitism | -173.33 | -282.33 | -238.56 |
| MizunoAIEV-Dora-Elitism | -142.11 | -255.33 | -260.55 |
| MizunoAIEV-Dora-NoElitism | -161.33 | -286.44 | 196.33 |

**Table 2.** Health difference between optimized agents and state-of-the-art bots. Differences between row bots against column ones.

## 4   Conclusions and future work

This work presents a preliminary study on the application of different schemes of Genetic Algorithms to optimize agents designed to combat in a 1 vs 1 fighting videogame. The results, even if they are not extraordinary, show some improvements over a very good starting agent.

Thus, we will continue this research line trying to improve GA optimization results by means of a higher explitation factor, more advanced schemes, a more suited configuration as well as specific and adapted operators (such as a better fitness function). We will also perform a more complete test against top-level agents from the Fighting AI Competition.